# VideoStory Embeddings Recognize Events when Examples are Scarce

Amirhossein Habibian, Thomas Mensink, Cees G. M. Snoek

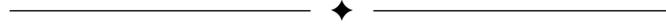

**Abstract**—This paper aims for event recognition when video examples are scarce or even completely absent. The key in such a challenging setting is a semantic video representation. Rather than building the representation from individual attribute detectors and their annotations, we propose to learn the entire representation from freely available web videos and their descriptions using an embedding between video features and term vectors. In our proposed embedding, which we call *VideoStory*, the correlations between the terms are utilized to learn a more effective representation by optimizing a joint objective balancing descriptiveness and predictability. We show how learning the VideoStory using a multimodal predictability loss, including appearance, motion and audio features, results in a better predictable representation. We also propose a variant of VideoStory to recognize an event in video from just the important terms in a text query by introducing a term sensitive descriptiveness loss. Our experiments on three challenging collections of web videos from the NIST TRECVID Multimedia Event Detection and Columbia Consumer Videos datasets demonstrate: i) the advantages of VideoStory over representations using attributes or alternative embeddings, ii) the benefit of fusing video modalities by an embedding over common strategies, iii) the complementarity of term sensitive descriptiveness and multimodal predictability for event recognition without examples. By it abilities to improve predictability upon any underlying video feature while at the same time maximizing semantic descriptiveness, VideoStory leads to state-of-the-art accuracy for both few- and zero-example recognition of events in video.

## 1 INTRODUCTION

This paper strives to recognize events such as *parking a vehicle*, *flash mob*, and *renovating a home* in web video content. A problem of increasing importance in a world that is swiftly adapting to video communication. The leading computer vision and multimedia retrieval solutions for this challenging problem, *e.g.,* [32], [50], [65], [71], [78], all learn to recognize events with the aid of many labeled video examples. However, as events become more and more specific, we consider it unrealistic to assume that ample examples to learn from will be commonly available. In practical use only a handful of video examples, an event name and an event definition may be present, such as the ones in Figure 1. We aim for event recognition when video examples are scarce or even completely absent.

Recognizing events from few or zero examples imposes constraints on the video representation. End-to-end deep learning of the video representation [27], [36], [68] demands too many video examples. In [36] for example, Karpathy *et al.* exploit more than 1 million YouTube videos and their sport category labels to learn the video representation. Alternatively, Xu *et al.* [78] show how a very deep convolutional neural network (CNN) [60] intended for image classification can be leveraged as representation for event recognition in video. They use responses from intermediate layers of the network to represent frames, which are aggregated over the video using VLAD encoding [26]. When combined with a linear SVM, excellent results on the leading NIST TRECVID event detection benchmarks [54] are reported for scenarios where many and few examples are available. The CNN video representation outperforms more traditional video encodings such as improved dense trajectories [71], [72] and multimedia representations combining appearance, motion and audio features [49], [52], [81]. However, both the learned and engineered representations are incapable, nor intended to, recognize events when examples are completely absent. We propose a video representation that can leverage any underlying feature, be it a CNN, improved dense trajectories and/or audio features, while being capable of few-example *and* zero-example event recognition with state-of-the-art accuracy.

The key to few- and zero-example event recognition is to add meaning to the video representation. Inspired by the success in image classification [16], [38], [55], many rely on the prediction scores made by a set of pre-trained attribute classifiers [4], [15], [21], [24], [31], [41]–[43], [45], [47], [79]. In [21] Habibian *et al.* study the properties of 1,346 attribute classifiers trained from ImageNet [58] and TRECVID [54] for representing and recognizing events in web video. Rather than pre-specifying and manually labeling each individual attribute in advance, the attributes can also be learned on top of imagery and (weak) labels harvested from the web [9], [11], [76], [80]. To assure visual predictability of the discovered attribute detectors a common tactic is to leverage part of the harvested data for validation [3]. However, a drawback is that many attribute labels rarely occur. For these infrequent labels only a limited number of positive examples are available, which leads to a biased estimation of their classification reliability. As a consequence, many of the discovered attributes might be overfitted to their small training set and do not generalize well for new videos. We also discover our semantic representation from the web, but rather than selecting individual, and often unreliable, classifiers per attribute label, we prefer to combine labels automatically into more predictable attributes. By combining labels, more training examples are available and a more

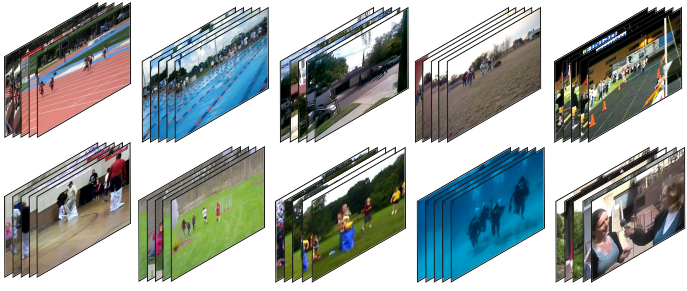

**Event Name:** `Winning a race without a vehicle`
**Definition:** `An individual (or more) succeeds in reaching a pre-deter-`
`mined destination before all other individuals, without vehicle as-`
`sistance or assistance of a horse or other animal. Racing generally`
`involves accomplishing a task in less time than other competitors.`
`The only type of racing considered relevant for the purposes of`
`this event is the type where the task is traveling to a destination,`
`completed by a person(s) without assistance of a vehicle or animal.`
`Different types of races involve different types of human ...`

Fig. 1. Video exemplars (top) and the textual definition (bottom) of the event *winning a race without a vehicle* to illustrate one of the events studied in this paper. Following the NIST TRECVID evaluation guidelines [54], the textual definition is for zero-example recognition, and the ten provided video exemplars are for few-example recognition of the event.

robust representation is obtained, without losing descriptive ability.

In this paper we present a semantic representation learning algorithm for videos. Instead of relying on pre-specified attribute labels, we learn the representation from freely available web videos and their descriptions. To this end, we propose an embedding between the video features and their textual descriptions, which is learned by utilizing the correlations between the words in the descriptions. We learn the embedding by minimizing a joint objective function balancing the descriptiveness and predictability of the learned video representation. Our proposed embedding is able to leverage the multiple modalities which coexist in video to learn a more reliable semantic representation. We call our video representation *VideoStory*, since it strives to encode the story of a video from its description.

A preliminary version of this article appeared as [20]. The current version adds *i)* representation learning for multiple video modalities, *ii)* representation learning for zero-example event recognition, *iii)* an additional TRECVID video data set, revised experiments, and improved baselines, all using a recent video CNN feature [64], [78], and *iv)* a new related work section, which will be discussed next.

## 2 RELATED WORK

In this section we focus on three directions of related work we deem most closely connected to ours.

### 2.1 Representations for Event Recognition

Until recently most event recognition methods exploit video representations based on densely extracted low-level visual features, such as HOG/HOF [39], or MBH [53], [72], often combined with audio features like MFCC features [34], [50], [65]. Currently, most methods extract frame-based deep convolutional neural network (CNN) features [37], [60], [64], using the responses from intermediate layers of the CNN, which is pre-trained on ImageNet images [58], see *e.g.,* [9], [31], [48], [78], [81]. To obtain per video descriptors the local/frame-based descriptors are aggregated by their mean, by using the Fisher vector [59] or VLAD encoding [26]. Despite the fact that these low-level based video representations obtain state-of-the-art event recognition performance, they suffer from two drawbacks. First, because of their high-dimensionality, training effective event classifiers on these representations require a sufficient number of training examples to prevent overfitting. Second, all these representations are incapable of providing a semantic interpretation of the video, which is crucial for zero-example recognition.

**Semantic Representations** To obtain a semantic representation for videos, inspiration is obtained from describing images with attributes [18], [38], and the video is represented by its attribute prediction scores. Creating the training data for a set of task-specific attribute classifiers manually involves lots of annotation effort, which is restrictive. Therefore, often public available datasets, such as ImageNet [58] and TRECVID [54], are used to train the attribute classifiers [21], [42], [45]. Although this overcomes the need for (additional) manual annotation, the attributes of these datasets are not necessarily descriptive for event recognition.

To tune the attributes for the task at hand, several works aim to automatically discover the attributes from web images/videos and their textual descriptions [9], [11], [76], [80]. For example Wu *et al.* [76], start by selecting the most frequent/relevant terms from a set of provided event descriptions as attributes. Then, inspired by Berg *et al.* [3], they use Internet search engines like Google and YouTube to gather positive examples. They assure visual predictability of the discovered attribute detectors by cross-validation. Similarly the work of Ye *et al.* [9] relies on the website WikiHow to obtain event descriptions, and the visual predictability of the selected terms is ensured by keeping only those terms which are present in existing image classification datasets. We refer to these methods as *term attributes* since they all discover the attributes from the terms in the descriptions.

Despite their effectiveness in obtaining training data, term attributes suffer from three drawbacks. First, descriptions have a long-tail distribution, therefore many terms rarely occur and these will not have enough positive examples to train reliable visual classifiers. Second, term attributes are selected mostly based on their visual prediction accuracy, while for effective event recognition the selected term attributes should also be descriptive for the target events. Third, contextual information is lost, since term attributes are learned independently by binary classifiers.

In this paper, we also learn the representation from videos and their descriptions. But, rather than selecting individual, and often unreliable, term attributes, we learn the entire representation by automatically combining the terms through embedding. In our proposed embedding, the correlation between the terms are utilized to learn a more effective representation, which is predictable *and* descriptive.

## 2.2 Embedding Videos and their Descriptions

To learn correspondences between the visual domain and textual descriptions different embedding methods have been proposed for various purposes *i.e.,* image annotation [5], image classification [2], [75], image captioning [35], [70], video to text translation [14], [77] and cross-modal retrieval [10], [17], [51].

Canonical correlation analysis (CCA) [23] is the classical unsupervised manner to relate different modalities and can be used for cross-modal retrieval [10]. CCA is the multimodal generalization of PCA, and can be computed as a generalized eigenvalue problem on the cross-covariance matrix between the visual and textual features. It finds a sequence of uncorrelated projections in which the cross-correlation between the modalities is maximized. This is not necessary suited to obtain a discriminative representation, as is also validated by our experiments.

Probabilistic topic models, such as the seminal *correspondence latent Dirichlet allocation* [5] and its extensions [13], [57], learn correspondences by extracting a set of correlated visual and textual topics from imagery and their captions. Despite their effectiveness for captioning images and videos [14], these methods are not designed to learn a representation for the purpose of recognition. Moreover, we note that by design these models are only applicable on discrete features, and therefore they cannot leverage the state-of-the-art video features used in our paper.

Embeddings for multi-class image classification, such as [2], [75] find a low-dimensional subspace in which multi-class classification is performed. The subspace is found jointly with the multi-class classifier by minimizing a classification loss [75], or a zero-shot classification loss [2]. In contrast to this multi-class image classification setting, we are interested in finding descriptions modeled as a multi-label video classification problem. Moreover, we argue that solely optimizing for classification is not sufficient to obtain a predictable and descriptive video embedding.

Recently deep neural network architectures have been proposed to learn multimodal correspondences for retrieval [17], [51] and captioning [35], [70]. Notably the multimodal recurrent neural networks capture the sequential ordering between the words in image captions and generate more accurate image captions, compared to the probabilistic topic models. However, training deep neural networks generally requires large amounts of training examples, while our purpose is to learn from few- and zero-examples.

## 2.3 Zero-Example Event Recognition

In zero-example event recognition the goal is to recognize an event, only based on a given textual *event definition*, without using any video examples. The event definition is usually provided in the form of a title and description, see Figure 1. This zero-example setting is beyond the conventional zero-shot image recognition of objects and scenes [16], [38], [56], where often a training set of related classes is available as well as pre-specified class-to-attribute mappings. Recently, this challenging event recognition problem has attracted a lot of attention because of its high practical value and the corresponding TRECVID benchmark task initiated by NIST [54]. The common approach for zero-example event recognition is to represent videos and the event queries using a semantic representation, and to rank all video representations based on the cosine similarity with the event query.

To represent the video mostly attributes and term-attributes [9], [11], [29], [30], [76], [80] are exploited. Extensions include combining attributes using logical operators [19], using video-segments to detect attributes [46], and adjustment of attribute scores using an ontology structure [31]. In addition to attributes, automatic speech recognition and optical character recognition have been considered to enrich the semantic video representations [12], [31], [76].

To answer the event query, usually term matching between the semantic video representation and the event definition is performed. The retrieval can be enriched by using contextual information, such as word embeddings and term co-occurrences [7], [80]. For the final ranking, the cosine similarity can be extended by pseudo relevance feedback mechanisms, such as self-paced ranking [29]. This has been used to improve the zero-shot event recognition by re-ranking [7], [30], [31]. Our representation learning is orthogonal to these efforts and can be joined with them to further improve the event recognition performance.

## 3 VIDEOSTORY EMBEDDINGS

Our goal is to learn a representation function $f : \mathcal{X} \to \mathcal{S}$, which maps each low-level video representation $x_i \in \mathcal{X}$ into the semantic representation $s_i \in \mathcal{S}$. The representation function is trained on a collection of videos and their semantic descriptions, which we use in the form of binary term vectors $y_i \in \mathcal{Y}$.

In the common attribute learning approach for learning the representation, the semantic representations are defined to be in the same space as the term vectors ($\mathcal{S} = \mathcal{Y}$). However, in practice, the term vectors are intrinsically noisy and highly sparse, which undermines their effectiveness as labels to train the representation function. Therefore, we propose to learn the semantic representation on a lower dimensional projection of the term vectors, which are less sparse and less noisy.

We formulate the representation function as an embedding, coined as VideoStory, which is learned by balancing two compelling forces in a joint optimization:

1) *Descriptiveness*, to preserve the information encoded in the video descriptions as much as possible, and
2) *Predictability*, to ensure that the representation could be effectively recognized from video content.

We first introduce the VideoStory embedding based on visual features (Section 3.1), then we generalize our embedding to fuse any multimedia feature (Section 3.2), and finally we gear the embedding towards zero-example recognition (Section 3.3). We summarize our notation conventions in Table 1.

### 3.1 Objective Function

We start from a dataset of videos, represented by video features $X$, and their textual descriptions, represented by binary term vectors $Y$, indicating which terms are present in

4TABLE 1
Summary of the core notation used for VideoStory.

| Notation | Description |
|---|---|
| $N$ | Number of videos |
| $M$ | Number of unique terms in descriptions |
| $D$ | Dimensionality of low-level feature |
| $J$ | Number of low-level features to fuse |
| $k$ | Dimensionality of VideoStory embedding |
| $\boldsymbol{X} \in \mathbb{R}^{D \times N}$ | Matrix of low-level video features |
| $\boldsymbol{Y} \in \{0,1\}^{M \times N}$ | Matrix of binary term vectors |
| $\boldsymbol{W} \in \mathbb{R}^{D \times k}$ | VideoStory visual projection |
| $\boldsymbol{A} \in \mathbb{R}^{M \times k}$ | VideoStory textual projection |
| $\boldsymbol{S} \in \mathbb{R}^{k \times N}$ | VideoStory embedding |
| $\boldsymbol{H} \in \mathbb{R}^{M \times M}$ | Diagonal matrix with per-term weights |
| $\boldsymbol{x}_i, \boldsymbol{y}_i, \boldsymbol{s}_i$ | The column representing the $i$-th video |

**Algorithm 1: Pseudocode for learning the VideoStory**

**input** : $\boldsymbol{X}, \boldsymbol{Y}, k, \eta$ (step-size), $m$ (max-epochs)
**output**: $\boldsymbol{W}$ and $\boldsymbol{A}$

$\boldsymbol{A}$, and $\boldsymbol{S} \leftarrow$ SVD decomposition of $\boldsymbol{Y}$
$\boldsymbol{W} \leftarrow$ random (zero-mean)
**for** $e \leftarrow 1$ **to** $m$ **do**
$\quad$ **for** $i \leftarrow 1$ **to** $N$ **do**
$\quad\quad$ Pick a random video-description pair $(\boldsymbol{x}_t, \boldsymbol{y}_t)$
$\quad\quad$ Compute gradients w.r.t. $\boldsymbol{A}, \boldsymbol{W}$ and $\boldsymbol{s}_t$
$\quad\quad$ Update parameters:
$\quad\quad\quad \boldsymbol{A} \quad \leftarrow \boldsymbol{A} - \eta_t \nabla_{\boldsymbol{A}} L_{\text{VS}}$ $\quad$ see Eq. (4)
$\quad\quad\quad \boldsymbol{W} \quad \leftarrow \boldsymbol{W} - \eta_t \nabla_{\boldsymbol{W}} L_{\text{VS}}$ $\quad$ see Eq. (5)
$\quad\quad\quad \boldsymbol{S} \quad \leftarrow \boldsymbol{s}_t - \eta_t \nabla_{\boldsymbol{s}_t} L_{\text{VS}}$ $\quad$ see Eq. (6)
$\quad$ **end**
**end**
**return:** $\boldsymbol{W}$ and $\boldsymbol{A}$

each video description. Then, our VideoStory representation is learned by minimizing:

$$L_{\text{VS}}(\boldsymbol{A}, \boldsymbol{W}) = \min_{\boldsymbol{S}} L_d(\boldsymbol{A}, \boldsymbol{S}) + L_p(\boldsymbol{S}, \boldsymbol{W}), \quad (1)$$

where $\boldsymbol{A}$ is the textual projection matrix, $\boldsymbol{W}$ is the visual projection matrix, and $\boldsymbol{S}$ is the VideoStory embedding. The loss function $L_d$ corresponds to our first objective for learning a descriptive VideoStory, and the loss function $L_p$ corresponds to our second objective for learning a predictable VideoStory. The VideoStory embedding $\boldsymbol{S}$ interconnects the two loss functions. To the best of our knowledge this joint embedding is novel.

**Descriptiveness.** For the $L_d$ function, we use a variant of regularized Latent Semantic Indexing (LSI) [73]. This objective minimizes the quadratic error between the original video descriptions $\boldsymbol{Y}$, and the reconstructed translations obtained from $\boldsymbol{A}$ and $\boldsymbol{S}$:

$$L_d(\boldsymbol{A}, \boldsymbol{S}) = \frac{1}{2} \sum_{i=1}^{N} \|\boldsymbol{y}_i - \boldsymbol{A}\boldsymbol{s}_i\|_2^2 + \lambda_a \Omega(\boldsymbol{A}) + \lambda_s \Psi(\boldsymbol{S}), \quad (2)$$

where $\Psi(\cdot)$ and $\Omega(\cdot)$ denote regularization functions, and $\lambda_a \geq 0$ and $\lambda_s \geq 0$ are regularizer coefficients. We use the squared Frobenius norm for regularization, which is the matrix variant of the $\ell_2$ regularizer, i.e., $\Omega(\boldsymbol{A}) = \frac{1}{2}\|\boldsymbol{A}\|_F^2 = \frac{1}{2}\sum_i \|\boldsymbol{a}_i\|_2^2 = \frac{1}{2}\sum_{ij} a_{ij}^2$, the sum of the squared matrix elements. Similarly for the VideoStory matrix $\Psi(\boldsymbol{S}) = \frac{1}{2}\|\boldsymbol{S}\|_F^2$.

The main difference with regularized Latent Semantic Indexing [73] is their $\ell_1$ regularizer, $\Omega(\boldsymbol{A}) = \sum_i \|\boldsymbol{a}_i\|_1$, which enforces sparsity in the textual projection $\boldsymbol{A}$. However, with our larger representation (typically we use a dimensionality of $k$ between 512 and 2,048 in our experiments compared to only $k = 20$ in [73]) and fewer number of unique terms (around 10K, compared to 100K), enforcing sparsity is not necessary for good performance.

Note that other textual embeddings can be formulated similar to Eq. (2), when appropriate regularization functions $\Omega(\cdot)$ and $\Psi(\cdot)$ are used. For example using $\Omega(\cdot) = \|\cdot\|_1$ enforces sparsity [22], [44], [73], or in the extreme case that $\boldsymbol{A}$ is constrained such that each column has a single non-zero value, the objective becomes very close to methods that select the best single term labels [3], and when $\boldsymbol{A}$ is enforced to preserve the taxonomical term relations, the objective resembles taxonomy embedding [69], [74].

**Predictability.** The $L_p$ function measures the occurred loss between the VideoStory $\boldsymbol{S}$ and the embedding of video features using $\boldsymbol{W}$. Since the VideoStory $\boldsymbol{S}$ is real valued, as opposed to a binary or multi-class encoding, we can not rely on standard classification losses such as the hinge-loss used in SVMs. Therefore, we define $L_p$ as a regularized regression, similar to ridge regression:

$$L_p(\boldsymbol{S}, \boldsymbol{W}) = \frac{1}{2} \sum_{i=1}^{N} \|\boldsymbol{s}_i - \boldsymbol{W}^\top \boldsymbol{x}_i\|_2^2 + \lambda_w \Theta(\boldsymbol{W}), \quad (3)$$

where we use (again) the Frobenius norm for regularization of the visual projection matrix $W$, $\Theta(\boldsymbol{W}) = \frac{1}{2}\|\boldsymbol{W}\|_F^2$, and $\lambda_w$ is the regularization coefficient.

**Joint optimization.** To handle large scale datasets and state-of-the-art high-dimensional visual features, e.g., Fisher vectors [59] on video features [72] or deep learned representations [37], we employ a Stochastic Gradient Descent (SGD) [6] optimization, summarized in Algorithm 1. The number of passes over the datasets (*epochs*) and the step-size $\eta$ are hyper-parameters of SGD.

The VideoStory objective function, as given in Eq. (1), is convex with respect to matrix $\boldsymbol{A}$ and $\boldsymbol{W}$ when the embedding $\boldsymbol{S}$ is fixed. In that case, the joint optimization is decoupled into Eq. (2) and Eq. (3), which are both reduced to a standard ridge regression for a fixed $\boldsymbol{S}$. Moreover, when both $\boldsymbol{A}$ and $\boldsymbol{W}$ are fixed, the objective in Eq. (1) is convex w.r.t. $\boldsymbol{S}$. Therefore we use standard SGD by computing the gradients of a sample w.r.t. the current value of the parameters, and we minimize $\boldsymbol{S}$ jointly with $\boldsymbol{A}$ and $\boldsymbol{W}$.

Lets denote a randomly sampled video and description pair at step $t$ by $(\boldsymbol{x}_t, \boldsymbol{y}_t)$, and let $\boldsymbol{s}_t$ denote the current VideoStory embedding of sample $t$. The gradients of Eq. (1) for this sample w.r.t. $\boldsymbol{A}, \boldsymbol{W}$ and $\boldsymbol{s}_t$ are given by:

$$\nabla_{\boldsymbol{A}} L_{\text{VS}} = -(\boldsymbol{y}_t - \boldsymbol{A}\boldsymbol{s}_t)\boldsymbol{s}_t^\top + \lambda_a \boldsymbol{A}, \quad (4)$$

$$\nabla_{\boldsymbol{W}} L_{\text{VS}} = -\boldsymbol{x}_t (\boldsymbol{s}_t - \boldsymbol{W}^\top \boldsymbol{x}_t)^\top + \lambda_w \boldsymbol{W}, \text{ and} \quad (5)$$

$$\nabla_{\boldsymbol{s}_t} L_{\text{VS}} = -\boldsymbol{A}^\top (\boldsymbol{y}_t - \boldsymbol{A}\boldsymbol{s}_t) + (\boldsymbol{s}_t - \boldsymbol{W}^\top \boldsymbol{x}_t) + \lambda_s \boldsymbol{s}_t. \quad (6)$$

The effect of joint learning the descriptiveness and the predictability, becomes clear in Eq. (6), where both the textual projection matrix $\boldsymbol{A}$ and visual projection matrix $\boldsymbol{W}$ contribute to learning the VideoStory embedding $\boldsymbol{S}$. This embedding $\boldsymbol{S}$ is subsequently used to obtain the textual projection $\boldsymbol{A}$ matrix, in Eq. (4), and the visual projection $\boldsymbol{W}$ matrix, in Eq. (5). This leads to the VideoStory embedding, which is both descriptive, by preserving the textual information, and predictable, by minimizing the visual prediction loss.

The parameters $A$, $S$, and $W$ can be initialized by random numbers with zero-mean. However, we experimentally observed that initializing the $A$ and $S$ matrices by singular value decomposition (SVD) of the term vectors $Y$, speeds up the convergence.

After training the visual and textual projection matrices, they are used to predict the VideoStory representation and the term vector of each video. In the case that both a video $x_i$ and description $y_i$ are given, we could obtain the VideoStory representation by returning $s_i$ from Eq. (1), while keeping both $A$ and $W$ fixed. However, in practice most videos are not provided with a description. Therefore, we use

$$s_i = W^\top x_i, \quad (7)$$

to predict our VideoStory representation from the low-level video features $x_i$. Moreover, using the predicted representation $s_i$, the term vectors for each unseen video are predicated as follows:

$$\hat{y}_i = A s_i, \quad (8)$$

where the terms with the highest values are most relevant for this video, see the illustration in Figure 2

### 3.2 VideoStory Fusion

Videos are inherently multimodal. In general any video contains appearance, motion, and audio cues and sometimes even textual information in the form of subtitles or speech recognition scripts. Fusing the different modalities is typically achieved by early-fusion, *i.e.*, fusion at the level of the representations, and late-fusion, *i.e.*, fusion at the level of prediction scores [61]. Both fusion strategies have been shown to be effective for understanding complex events as well, *e.g.*, [34], [50], [67], [76]. We propose *VideoStory*$^\mathcal{F}$, which extends the VideoStory embedding by learning the semantic representation from multiple modalities.

A straightforward approach to learn the multimodal semantic representation is by fusing multiple VideoStory embeddings, which are independently trained per modality. More specifically, given the video descriptions and low-level features from one modality, a VideoStory embedding is trained as detailed in Algorithm 1. Then the learned embeddings from all modalities are fused to obtain the multimodal representation. Despite its simplicity, training the VideoStory embeddings per modality is not optimal as it ignores the interrelations between the various modalities for learning the embeddings.

We aim for learning the embeddings jointly over all the video modalities. Our intuition is that the semantic representation is more effective if it is predictable from *all* the modalities rather than from each individual modality. For *VideoStory*$^\mathcal{F}$ we adjust the predictability loss, of Eq. (3), to incorporate a weighted combination of the predictability from all $J$ modalities as follows:

$$L_p^\mathcal{F}(S, \mathcal{W}) = \sum_{j=1}^{J} L_p(S, W^j) \quad (9)$$

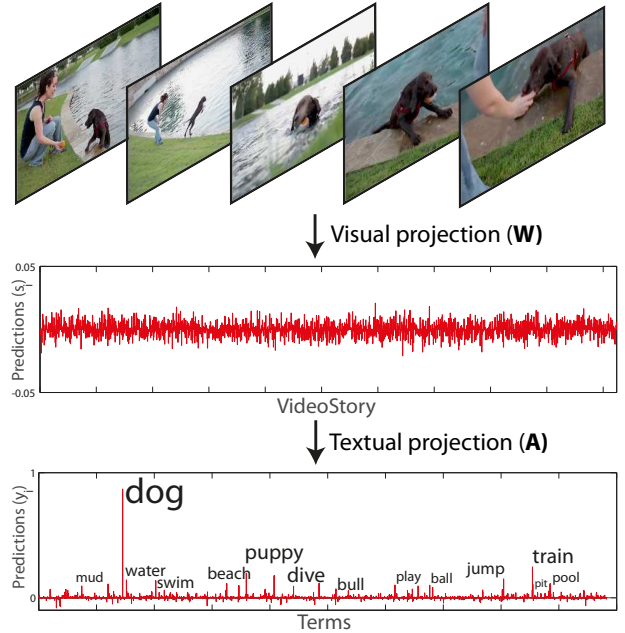

Fig. 2. VideoStory prediction: From the low-level video features the VideoStory representation and the term vector are predicted using the visual projection matrix $W$ and the textual projection matrix $A$.

where $S$ is the multimodal VideoStory embedding, and $\mathcal{W} = \{W^j, j = 1\ldots J\}$ is a set of projection matrices from the $J$ modalities. Each feature projection matrix $W^j \in \mathbb{R}^{D_j \times k}$ projects the low-level feature $x_i^j \in \mathbb{R}^{1 \times D_j}$ extracted from the video into the VideoStory representation $s_i$. Note that, instead of weighing each modality equally, a term $\gamma_j \geq 0$ could be used to weight the importance of each modality, if sufficient training examples are available for cross-validation of these weighting parameters.

The objective function Eq. (1) is still convex with respect to the parameters $S$, $A$, and $W^j$ when the other parameters are fixed. However, the gradient with respect to $s_t$, Eq. (6) becomes:

$$\nabla_{s_t} L_{\text{VS}} = -A^\top (y_t - A s_t) + \left(s_t - \sum_j W^{j\top} x_t^j\right) + \lambda_s s_t. \quad (10)$$

It can be seen that all the modalities are jointly contributing to learn the multimodal VideoStory representations $S$.

After training the textual and feature projection matrices, they are used to extract the multimodal VideoStory representation. Each feature projection matrix $W^j$ predicts the VideoStory representation based on its underlying modality as follows:

$$s_i^j = W^{j\top} x_i^j. \quad (11)$$

The final VideoStory representation is the concatenation of the per modality embeddings: $s_i = [s_i^1, \ldots, s_i^J]$.

### 3.3 VideoStory for Zero-Example Event Recognition

The objective function for learning the VideoStory embedding for few-example recognition is not necessarily optimal for the zero-example setting. In Section 3.1, the descriptiveness loss $L_d$ is defined as the overall error in reconstructing *all* the terms from the VideoStory representations, see Eq. (2). With this definition, the descriptiveness loss is biased toward the more frequent terms, as minimizing their reconstruction error leads to a higher decrease in the overall error. Consequently, the terms which are infrequent in the descriptions might be discarded, which degrades their prediction accuracy from video features. This undermines the effectiveness of our representation learning for zero-example setting, where the accuracy in predicting the terms from video is crucial.

We extend the VideoStory embedding to learn a video representation, which is effective for zero-example event recognition. Our extension, which we coin as *VideoStory*$_0$, minimizes the reconstruction error of the terms with respect to their importance for describing the events, rather than their frequency in the VideoStory train data. For this purpose, we use a *term sensitive* descriptiveness loss, as follows:

$$L_d^{ts}(\boldsymbol{A}, \boldsymbol{S}) = \frac{1}{2}\sum_{i=1}^{N}\|\boldsymbol{H}^{\frac{1}{2}}\left(\boldsymbol{y}_i - \boldsymbol{A}\boldsymbol{s}_i\right)\|_2^2 + \lambda_a \Omega(\boldsymbol{A}) + \lambda_s \Psi(\boldsymbol{S}), \quad (12)$$

where $\boldsymbol{H} \in \mathbb{R}^{M \times M}$ is a diagonal matrix, denoting the importance of each term for describing the events. By setting a relatively high value for $h_{jj}$ for term $j$, its reconstruction error is more penalized compared to the other terms. Hence the term is expected to be more precisely reconstructed.

We determine the term importance matrix $\boldsymbol{H}$ by relying on the presence/absance of terms in the textual event definitions. Our assumption is that the terms, which are present in event definitions are more important than the absent terms. More precisely, we set each element of the importance matrix $h_{jj}$ to $\alpha$, if the term $j$ is present, and $1-\alpha$ if the term $j$ is absent in the event definitions. $\alpha$ is a balancing parameter between 0 and 1, which should be higher than 0.5 to assign more importance to the present terms. We empirically set this parameter to 0.75 in all our zero-example event recognition experiments. The importance matrix can be extracted either for all events jointly or separately per event. We opt for the latter in our experiments, *i.e.*, extracting $\boldsymbol{H}^e$ based on the description for event $e$, since in the literature the events are usually treated separately.

After training the visual and textual projection matrices, we follow the standard pipeline for zero-example event recognition: First, each test video is represented by predicting its term vector $\hat{\boldsymbol{y}}_i^e$ based on Eq. (8), using the representation learned using $H^e$. Second, we translate the textual event definition into the event query, denoted as $\boldsymbol{y}^e \in \mathbb{R}^M$, by matching the terms in the event definition with the $M$ unique terms in the VideoStory training data. Finally, the ranking is obtained by measuring the similarity between the video representations and the event query based on the cosine similarity:

$$s_e(\boldsymbol{x}_i) = \frac{\boldsymbol{y}^{e\top}\hat{\boldsymbol{y}}_i^e}{||\boldsymbol{y}^e||\;\;||\hat{\boldsymbol{y}}_i^e||}. \quad (13)$$

**Multimodal fusion.** Finally, we can leverage the multimodal features for zero-example event recognition by combining the multimodal predictability loss and the term sensitive descriptiveness in a joint objective:

$$L_{\text{VS}}(\boldsymbol{A}, \boldsymbol{W}) = \min_{\boldsymbol{S}} L_d^{ts}(\boldsymbol{A}, \boldsymbol{S}) + L_p^{\mathcal{F}}(\boldsymbol{S}, \boldsymbol{W}). \quad (14)$$

We coin the learned video representation *VideoStory*$_0^{\mathcal{F}}$, since it is learned on multiple video modalities for zero-example event recognition.

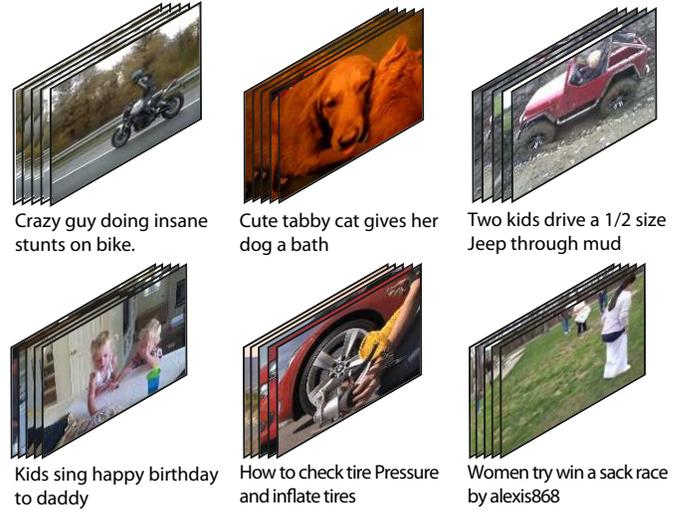

Fig. 3. Example videos and title captions from the VideoStory46K dataset [20], which we use for VideoStory representation learning.

## 4 EXPERIMENTAL SETUP

### 4.1 Datasets

We first introduce the dataset used for learning the VideoStory embeddings. Then, we detail the video datasets by which we evaluate the event recognition experiments.

#### 4.1.1 VideoStory Learning

In all the experiments, we learn the VideoStory embeddings on the ***VideoStory46K*** [20] dataset. This collection encompass 45,826 videos harvested from YouTube, with a total length of 743 hours. Every video comes with a short title caption provided by the user who has uploaded the video, as shown in Figure 3. There are 19,159 unique terms in the captions, most of them occurring infrequently. We filter out the terms occurring only once as they generally are misspelled terms, numbers, or noisy terms. It provides us with 9,828 unique terms, which are used in our experiments.

#### 4.1.2 Event Recognition Evaluation

We perform our event recognition experiments on the challenging TRECVID Multimedia Event Detection corpus [62] and the Columbia Consumer Video collection [33]. These corpora contain more than 42K videos in total, including user generated web videos with a large variation in quality, length and content.

**TRECVID Multimedia Event Detection (MED) [62].** This dataset is introduced by NIST as a benchmark for event recognition. We perform our experiments on the two latest releases of the dataset, referred to as ***MED 2013*** and ***MED 2014***. Each dataset includes videos from 20 complex events categories, but with 10 events overlapping as listed in Table 5. Each dataset includes three labeled video partitions: Event Kit training, Background training, and test set MED including 200, 5K, and 27K videos, respectively[1]. Apart from the videos, a textual definition is provided per event, which explicates the event as unformatted plain text, such as the one shown in Figure 1.

---

1. There is also a PROGRESS set with 98K videos, but this partition is for blind testing by NIST only.

We perform our few-example and zero-example event recognition experiments by exactly following the *10Ex* and *0Ex* evaluation procedure outlined by the NIST TRECVID event recognition task [54]. It means that, in the few-example event recognition experiments, training data for each event is composed of 10 positive videos from the Event Kit training data along with about 5K negative videos from the Background training data. The event recognition results for each event classifier are reported on the 27K videos from test set MED. In our zero-example event recognition experiments, we rely on the provided textual event definitions to create an event query vector. Then the event recognition performances are reported on the test set MED.

**Columbia Consumer Video (CCV) [33].** This dataset contains 9,317 user-generated videos from YouTube including over 210 hours of videos in total. The dataset contains ground truth annotations at video-level for 20 semantic categories, *i.e.*, *wedding reception* and *music performance*. We use the standard partitioning of the dataset, but we use only 10 positive examples per event in the training data. These 10 are selected based on alphabetical order of the respective video names, we ignore the remaining positive examples in the train set. We report event recognition results on the standard test partition. We denote our redefinition of the CCV dataset for few-example event recognition as $CCV_{10}$.

### 4.2 Video Features

To cancel out accidental effects of the choice for the underlying features, we consider the same set of appearance, motion and audio features for all our experiments, the various baselines, and our VideoStory variants. The appearance features are used for all the experiments. The motion and audio features are added for the experiments involving the multimodal fusion.

**Appearance.** We adopt the *video CNN* representation [78] as appearance features for event recognition, but found the very deep network of [64] to perform slightly better than [60]. For each video the frames are extracted by uniformly sampling every two seconds. Then, the CNN descriptors are extracted per frame as the 1K dimensional responses from the last fully connected layer (pool$_5$) of the Google Inception network [64]. We train the network on the 15,293 ImageNet categories with more than 200 examples, using the Caffe toolbox [28]. The final video CNN features are obtained by aggregating the frame descriptors over each video by VLAD encoding [26] with a codebook size of 20, resulting in a 20,480 dimensional vector.

**Motion.** We use the MBH descriptors along the motion trajectories [72] as motion features. The extracted 288-dimensional descriptors are reduced to 128 dimensions using PCA and are then aggregated per video using a Fisher vector [59], with 128 Gaussians resulting in a 32,768 dimensional vector. Each Fisher vector is power normalized, with $\alpha = 0.2$, as in [25].

**Audio.** We extract MFCCs descriptors [1] over a 10ms window. The descriptors consist of 13 values, 30 coefficients and the log-energy, along with their derivatives and the second derivatives. The MFCC descriptors are aggregated by Fisher vectors using a Gaussian Mixture Model with 256 components, resulting in a 46,080 dimensional vector.

### 4.3 Implementation Details

We learn the VideoStory embeddings by using 75% of the VideoStory46K dataset for training and 25% for validation to set the hyper-parameters of our model ($\lambda_w, \lambda_a, \lambda_s$) and of SGD (number of epochs, $\eta$). As the validation criterion we rely on the objective function value, when using $S = W^\top X$.

For few-example event recognition, the event classifiers are trained as binary SVM with RBF kernels, which are shown to be effective for semantic representations [47]. Following [78], [81] we set the SVM regularization and the RBF kernel parameters by a default value of 1, as the train set is not big enough to reliably estimate them by cross-validation.

As evaluation criteria we follow the standard convention in the literature [33], [54] by relying on the average precision (AP) per event, and we report the mean average precision (mAP) for overall accuracy.

## 5 EXPERIMENTS

### 5.1 VideoStory for Few-Shot Event Recognition

#### 5.1.1 Effect of Embedding

We first quantify the impact of embeddings for learning the representation by comparing VideoStory with term attribute baselines. The baselines learn the representations directly from terms without any embedding. We evaluate all video representations for few-example event recognition using a dimensionality varying from 32 to 8,192.

*1. Term attributes.* This representation is extracted by following the tradition of predicting relevant individual terms from the video descriptions *e.g.,* [9], [11], [76], [80]. A linear SVM classifier is trained per term. The classifiers which have the highest prediction accuracy, based on a 2-fold cross-validation [3], are selected as term attributes.

*2. Term attributes-f.* This baseline is similar to the previous baseline, but rather than using cross-validation to select the term attributes it simply selects the terms with the highest *frequency* in the descriptions.

**Results.** The results in Figure 4 demonstrate that VideoStory embeddings outperform the term attribute and term attribute-f baselines on all the three test sets.

Term attributes, which relies on the estimated reliability of individual term classifiers, performs worst. This representation suffers from two drawbacks. First, many of the visual terms are very specific and therefore incapable of characterizing the events of interest *i.e.,* `necklace`, `suitcase`, and `earring`. Although these terms can be accurately predicted from videos, they are incapable of providing a characteristic representation for event recognition. Second, many of the terms rarely occur in video descriptions. Hence, only a limited number of positive examples are available to learn the term classifiers, which leads to a biased estimation of their reliability. Consequently, many of the discovered visual terms overfit to their small training set and do not generalize well for new videos.

The drawbacks of term attributes are relaxed by simply relying on the most frequent terms. We observe the most frequent terms usually refer to characteristic attributes of events which are frequently used by humans when describing a video, *i.e.,* `car`, `girl`, and `kid`. Moreover, because of their large number of positive examples, the trained visual classifiers from the term attributes-f baseline are in general more reliable.

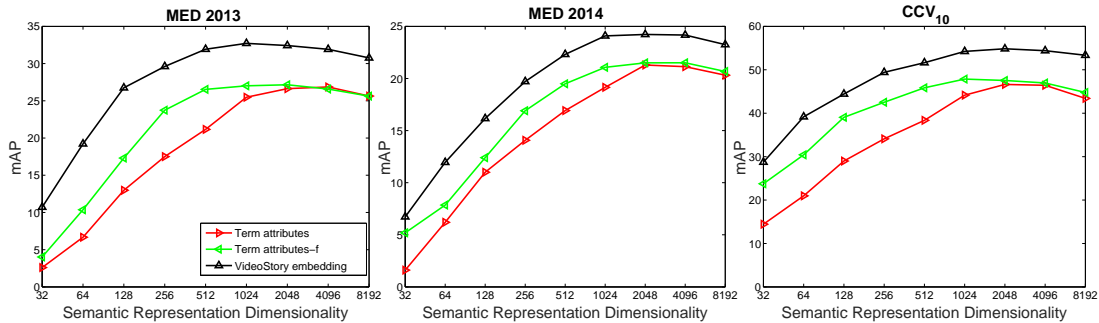

Fig. 4. Effect of embedding. The VideoStory embedding outperforms the term attribute and term attribute-f baselines, which are directly learned from the terms without embedding.

The VideoStory embedding represents the terms in a reduced-dimensional space, where correlated terms are usually combined together. Combining correlated terms leads to less correlation between dimensions of the learned representation. Moreover, as the positive examples for all correlated terms are combined, it provides more positive video examples to train visual classifiers, often leading to better accuracy.

We also compare the learning time for VideoStory versus the term-attribute baselines on a CPU Intel Xeon E5-2690@2.90GHZ with 256GB of memory. As Figure 5 shows, learning the VideoStory embedding is more efficient than both baselines. The term attributes baseline is least efficient as it trains individual classifiers for all the 9,828 terms within the collection, before selecting the most accurate ones. Term attributes-f is more efficient, as it only trains the classifiers for the most frequent terms. Our proposed VideoStory embedding has the fastest learning time as it learns the visual embeddings for the combination of terms rather than individual terms.

Besides its effectiveness for few-example event recognition, the VideoStory embedding also improves the representation learning time by reducing the term redundancies.

### 5.1.2 VideoStory vs Other Embeddings

We compare the effectiveness of our proposed VideoStory embedding with two alternative embeddings:

*1. CCA embedding.* This baseline learns the textual and visual projections by CCA [10], which maximizes the cross-correlation between the video features and descriptions. We experimentally observed that the embedding is even more effective when we PCA-reduce the video features to a dimensionality of 1,024 before learning the CCA embedding.

*2. Description embedding.* Similar to the VideoStory, this embedding is learned by minimizing the descriptiveness and predictability losses, but in two disjoint steps: The textual projection is first learned based on the regularized Latent Semantic Indexing [73], as in Eq. (2). Then the visual projection is learned separately, by minimizing the error for predicting the embedded descriptions from the video features based on ridge regression, as in Eq. (3).

**Results.** The results in Figure 6 demonstrate that the VideoStory embedding outperforms the CCA and the Description embedding baselines on all the three test sets for a dimensionality larger than 256.

We explain the gain over CCA by the fact that CCA is a symmetric embedding, it learns both the textual and visual projections in the same way. However, the textual and visual features have different distributions and properties, which may require different objective functions to learn their projections. This is achieved by our proposed VideoStory embedding as it relies on two separate LSI and ridge regression loss functions to learn the textual and visual projections, respectively.

We explain the improvement over the Description embedding by the fact that combining the terms based on textual correlation only does not necessarily imply that the corresponding video is visually correlated as well. As it happens, the terms `puppy` and `kid` have a high correlation in the descriptions but are visually dissimilar. Combining these two terms together, as is done by Description embedding, undermines the accuracy of the classifiers predicting them from videos. In contrast, in a VideoStory the correlated terms are combined only if their combination improves their classifier prediction. It prevents the combination of correlated terms which are visually dissimilar.

From now on we use a fixed 2,048 dimensional VideoStory representation, which is optimal based on Figure 4.

### 5.1.3 VideoStory vs other Representations

We evaluate the VideoStory by comparing it with state-of-the-art video representations for few-example event recognition:

*1. Low-Level.* In this baseline, the event classifiers are trained directly on the low-level video representations, without extracting a semantic representation. We rely on the video CNN features of [78], based on our implementation as detailed in Section 4.2.

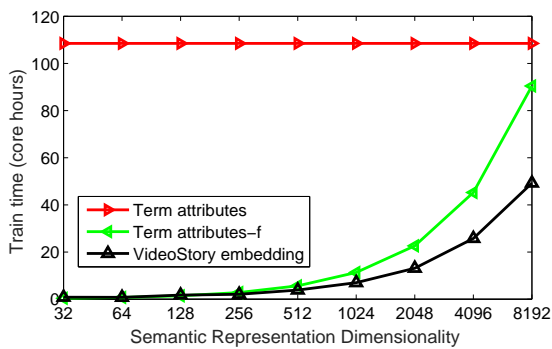

Fig. 5. Effect of embedding on computation efficiency. Learning the VideoStory embedding is more efficient than the term attribute and term attribute-f baselines, as it learns one classifier per group of terms rather than individual classifiers per term.

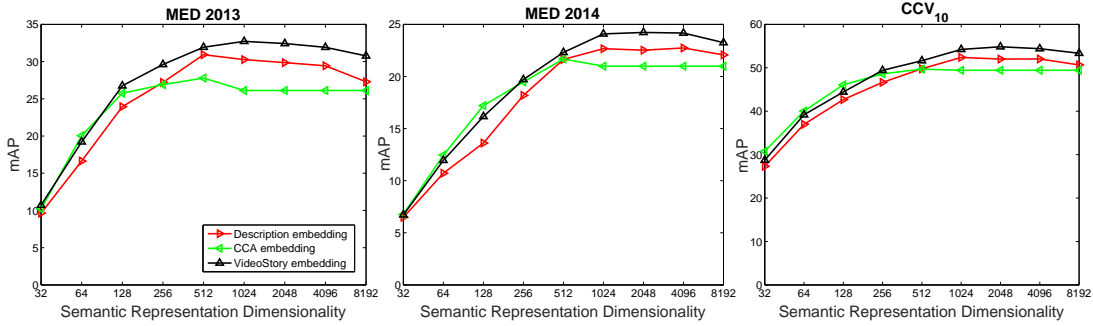

Fig. 6. VideoStory vs Other Embeddings. VideoStory embedding outperforms the CCA and the Description embedding baselines on all three test sets. The Description embedding is the closest competitor, but it suffers from embedding correlated terms which are visually dissimilar. CCA, uses the same objective function to learn the visual and textual embeddings, which is suboptimal due to intrinsic differences between the visual and textual features.

TABLE 2
VideoStory vs other Representations for few-example event recognition. Our proposed VideoStory outperforms the alternatives on all three test sets.

|  | Low-Level | Attributes | VideoStory |
| --- | --- | --- | --- |
| **MED 2013** | 28.1 | 22.5 | **32.4** |
| **MED 2014** | 21.8 | 17.2 | **24.2** |
| **CCV$_{10}$** | 50.0 | 48.8 | **54.8** |

*2. Attributes.* This representation is obtained by adopting the public ImageNet dataset as the source for training attribute classifiers as proposed in [21]. However, instead of training SVM classifiers on bag-of-words encoding of color SIFT descriptors [21], we upgrade the attributes by training a deep CNN with Google Inception architecture on the 15,293 ImageNet categories as detailed in Section 4.2.

**Results.** Table 2 shows that VideoStory outperforms the state-of-the-art attributes and low-level video representations on all three test sets.

By comparing the VideoStory and the attribute representation we observe a higher event recognition accuracy of 32.4 vs 22.5 for the MED 2013, 17.2 vs 21.8 for the MED 2014, and 54.8 vs 48.8 for the CCV$_{10}$ test set. We explain it by the fact that attribute baseline relies on the ImageNet categories as attributes. However, many of these pre-specified categories are not semantically relevant for the events of interest. For example, many of the ImageNet categories are devoted to specific animal species, which are not generally characteristics of the events. In contrast, the VideoStory embedding is automatically derived from the VideoStory46K dataset, which includes many descriptions relevant to events. Note, in contrast to previous work, our low-level representation outperforms the attribute representation, indicating the strong low-level features used in this work.

The results further demonstrate that the VideoStory outperforms the low-level representation. As an explanation, we speculate that the low-level representation is prone to overfitting due to its high dimensionality. More specifically, in the low-level baseline, the event classifiers are trained on the 20,480 dimensional low-level features only from 10 positive exemplars. This may lead to overfitting as a result of the curse of dimensionality. In contrast, the VideoStory representation transfers the semantics from descriptions into the video representation to alleviate the overfitting as a sort of regularization.

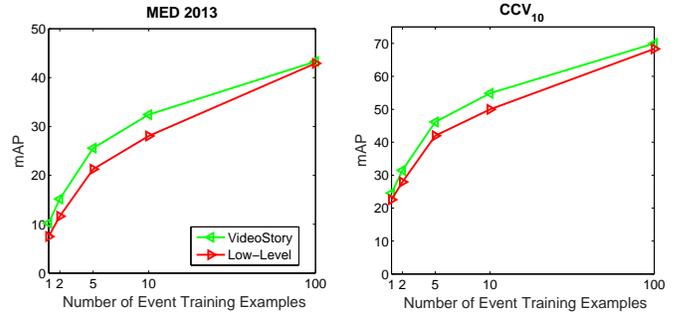

Fig. 7. From 1 to 100 examples. When the number of event exemplars are limited the VideoStory embedding outperforms the low-level representation. By increasing the number of training examples their difference becomes more subtle.

**From 1 to 100 examples.** For further investigation, we gradually increase the number of positive examples from 1 to 100, and evaluate the accuracy of the event classifiers trained on both the low-level and VideoStory representation. The positive examples are selected randomly[2] and the results are reported by averaging over 10 repetitions to compensate for the random effect. As Figure 7 shows, when the number of event train examples is less than 10, the VideoStory representation outperforms the low-level representation. However, by increasing the number of training examples to 100, the difference in the performances becomes more subtle, which confirms the hypothesis.

### 5.2 VideoStory Fusion

We evaluate the effectiveness of the VideoStory fusion (VideoStory$^{\mathcal{F}}$) proposed in Section 3.2 for few-example event recognition. We perform the experiments using the appearance, motion, and audio modalities, as detailed in Section 4.2.

#### 5.2.1 Effect of fusion

We study the impact of fusing multiple modalities for learning the video representation. We start from using only the appearance features (*Appearance*) and gradually add the motion (*+ Motion*) and audio features (*+ Audio*) to learn the video representation by our proposed VideoStory$^\mathcal{F}$.

---

2. The additional event exemplars are selected from the 100Ex evaluation procedure provided in the MED dataset

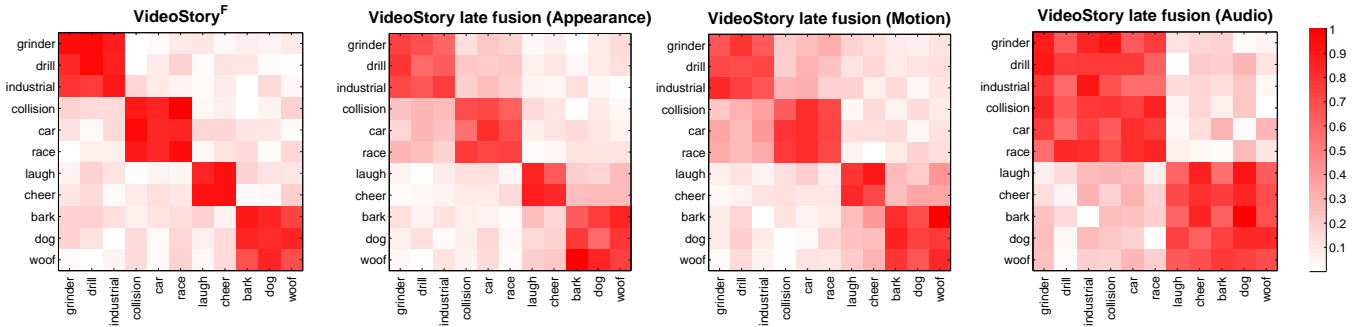

Fig. 8. Effect of learning the embeddings *jointly* over all the modalities by comparing VideoStory$^\mathcal{F}$ with the VideoStory late fusion baseline. Each textual projection matrix is visualized by plotting $\boldsymbol{A} \times \boldsymbol{A}^\top$, which reveals the learned term combinations. The VideoStory$^\mathcal{F}$ learn a more reasonable combination of terms compared to the VideoStory late fusion, where the embeddings are learned disjointedly per modality.

TABLE 3
Effect of fusion. Fusing more modalities to learn VideoStory$^\mathcal{F}$ leads to a more effective semantic representation.

| | **MED 2013** | | | | **MED 2014** | | | | **CCV$_{10}$** | | |
|---|---|---|---|---|---|---|---|---|---|---|---|
| *Event* | Appearance | + Motion | + Audio | *Event* | Appearance | + Motion | + Audio | *Event* | Appearance | + Motion | + Audio |
| Birthday party | 37.1 | 38.8 | **43.4** | Attempting a bike trick | 9.0 | **9.4** | 8.8 | Basketball | 65.6 | 67.9 | **68.6** |
| Changing a vehicle tire | 64.7 | 65.5 | **67.2** | Cleaning an appliance | 11.1 | 13.7 | **16.0** | Baseball | 58.6 | **59.6** | 59.5 |
| Flash mob gathering | 55.3 | 63.8 | **65.0** | Dog show | 83.7 | 88.0 | **89.7** | Soccer | 58.7 | 64.5 | **64.7** |
| Getting a vehicle unstuck | 59.3 | 65.2 | **65.3** | Giving directions to a location | 0.5 | **1.4** | 1.0 | Ice Skating | 70.2 | 74.5 | **75.1** |
| Grooming an animal | 24.3 | **29.4** | 28.2 | Marriage proposal | 0.3 | 0.5 | **0.6** | Skiing | 79.4 | 81.8 | **81.9** |
| Making a sandwich | 16.7 | 18.9 | **21.3** | Renovating a home | 11.5 | 11.9 | **12.9** | Swimming | 78.9 | **81.8** | 81.7 |
| Parade | 33.9 | **44.7** | 44.0 | Rock climbing | **13.8** | 14.2 | 13.7 | Biking | 67.3 | 69.0 | **70.2** |
| Parkour | 61.3 | **72.8** | 72.2 | Town hall meeting | 40.1 | 36.3 | **42.4** | Cat | 66.2 | 68.3 | **70.8** |
| Repairing an appliance | 44.5 | 49.7 | **57.4** | Winning a race without a vehicle | 22.1 | **26.8** | 26.7 | Dog | 65.4 | **68.9** | 68.6 |
| Working on a sewing project | 47.2 | **48.0** | 46.0 | Working on a metal crafts | 15.0 | 17.9 | **20.6** | Bird | 33.9 | 35.9 | **36.4** |
| Attempting a bike trick | 8.8 | **9.3** | 8.9 | Beekeeping | **54.1** | 53.5 | 48.1 | Graduation | 25.3 | **27.6** | 25.1 |
| Cleaning an appliance | 10.5 | 13.2 | **15.5** | Wedding shower | 19.7 | 31.5 | **40.3** | Birthday | 54.2 | 56.2 | **61.9** |
| Dog show | 81.6 | 84.9 | **85.9** | Non-motorized vehicle repair | 65.3 | 66.0 | **66.5** | Wedding Reception | 17.3 | 18.3 | **19.5** |
| Giving directions to a location | 0.6 | **1.0** | 0.9 | Fixing musical instrument | 25.4 | 31.0 | **44.0** | Wedding Ceremony | 45.6 | 53.5 | **58.4** |
| Marriage proposal | 0.3 | 0.4 | **0.5** | Horse riding competition | **40.5** | 37.9 | 35.6 | Wedding Dance | 51.3 | 58.6 | **61.7** |
| Renovating a home | 11.4 | 12.1 | **13.8** | Felling a tree | 12.4 | 16.2 | **22.3** | Music Performance | 41.0 | 41.6 | **51.1** |
| Rock climbing | 13.7 | **14.9** | 14.4 | Parking a vehicle | 17.3 | 19.5 | **21.9** | NonMusic Performance | 30.6 | 35.6 | **36.4** |
| Town hall meeting | 40.2 | 38.9 | **44.8** | Playing fetch | 1.3 | 1.3 | **1.4** | Parade | 49.4 | 64.0 | **64.0** |
| Winning a race without a vehicle | 21.9 | **28.8** | 27.8 | Tailgating | 32.9 | 34.5 | **37.7** | Beach | **74.5** | 73.9 | 72.9 |
| Working on a metal crafts | 15.2 | 18.0 | **20.4** | Tuning musical instrument | 8.7 | 8.3 | **15.1** | Playground | **63.4** | 61.9 | 58.1 |
| *mAP* | 32.4 | 35.9 | **37.1** | *mAP* | 24.2 | 26.0 | **28.3** | *mAP* | 54.8 | 58.2 | **59.3** |

TABLE 4
Comparison with other fusions. The VideoStory$^\mathcal{F}$ outperforms the alternative fusion tactics.

| | Early fusion | Late fusion | VideoStory early fusion | VideoStory late fusion | VideoStory$^\mathcal{F}$ |
|---|---|---|---|---|---|
| **MED 2013** | 32.6 | 33.8 | 33.8 | 33.6 | **37.1** |
| **MED 2014** | 27.0 | 27.0 | 27.0 | 26.1 | **28.3** |
| **CCV$_{10}$** | 56.3 | 55.8 | 56.9 | 55.4 | **59.3** |

**Results.** Table 3 shows that the more modalities are fused, the more effective video representation is learned by VideoStory$^\mathcal{F}$. It confirms that incorporating more modalities for estimating the predictability loss leads to a more effective representation learning.

By looking into the results, we observe that some events are benefiting more from adding the auditory modality, *i.e.*, *birthday party* and *music performance*. For these events, there are some distinctive terms which are more effectively predicted from the audio features, *i.e.*, `singing`, `clapping`, and `piano`. Similarly, some other events are more improved by including motion features, *i.e.*, *parkour* and *parade*, for which some distinctive terms such as `jumping`, `rolling`, and `running` are well predictable from the motion features.

It demonstrates that different modalities are complementary for predicting the terms, so they are all required to effectively predict the video descriptions.

### 5.2.2 Comparison with other fusions

We evaluate our proposed VideoStory$^\mathcal{F}$ by comparing it with the following fusion baselines:

*1. Early fusion.* This baseline fuses the modalities by simply concatenating the low-level features from all the three modalities into a longer feature vector. Then the event classifiers are trained and applied on the concatenated low-level features. This simple baseline is shown to be very competitive to the more complicated multiple kernel learning [67].

*2. Late fusion.* This baseline fuses the modalities at the level of event classification scores. A separate event classifier is learned by training an SVM on low-level video features per modality. For each test video, the final event detection score is obtained by averaging the detection scores predicted by each classifier, as evaluated in [50].

*3. VideoStory early fusion.* This baseline learns one VideoStory embedding on the concatenation of various low-level video features, using the standard VideoStory objective of Eq. (1). For each video, a semantic representation is extracted by applying the learned feature projection on the concatenation of the low-level video features. Then, the event classifiers are trained and applied on the semantic video representations.

*4. VideoStory late fusion.* This baseline learns three separate VideoStory embeddings, one per modality, using the standard VideoStory objective of Eq. (1). For each video, a semantic representation is obtained by concatenating the three VideoStory representations, which are predicted from each modality. Then, the event classifiers are trained and applied on the semantic video representations.

**Results.** The results are reported in Table 4. The VideoStory$^\mathcal{F}$ outperforms the alternative fusion baselines on all the three test sets. We explain the better performance of VideoStory$^\mathcal{F}$ compared to the early fusion and late fusion by the fact that both baselines train the event classifiers directly from low-level video features. However, in VideoStory$^\mathcal{F}$ the event classifiers are trained on the semantic VideoStory representations, which are more effective than the low-level features in general, as shown in Section 5.1.3 for the appearance features.

In the VideoStory late fusion the embeddings are learned *separately* per modality. In contrast, VideoStory$^\mathcal{F}$ relies on all the video modalities jointly to estimate the predictability loss, which in general is more reliable than the per modality predictability estimations. For further investigation, we visualizing the learned textual projection matrices $A$ in Figure 8. As shown in this figure, the VideoStory$^\mathcal{F}$ prevents some undesirable combination of terms that happen when the predictability losses are estimated separately per modality. For example, in the VideoStory which is learned only on audio features, the terms `laugh`, `cheer`, `bark`, `dog`, and `woof` are all combined as they have similar auditory features (see the right plot). However, the VideoStory$^\mathcal{F}$ does not combine the `laugh` and `cheer` with the `bark`, `dog`, and `woof` terms, as these terms are different in the appearance and motion features. Hence, the VideoStory$^\mathcal{F}$ learns a more reliable combination of terms, which leads to a more effective video representation.

VideoStory early fusion learns a single feature projection matrix on the concatenated video features. However, the low-level features from different modalities have a different intrinsic dimensionality, distribution, and meaning, which aggravates the learning from their concatenation. In contrast, the VideoStory$^\mathcal{F}$ learns separate feature projection matrices per modality, where each feature projection is optimized based on the features from one modality. It alleviates learning the feature projections, which leads to a more effective video representation.

To conclude, the VideoStory$^\mathcal{F}$ effectively fuses the features from various video modalities by embedding them into a mutual semantic representation learned jointly over all the modalities.

### 5.3 VideoStory for Zero-Example Recognition

We evaluate the effectiveness of the VideoStory$_0$ and its multimodal fusion VideoStory$_0^\mathcal{F}$, proposed in Section 3.3, for zero-example event recognition by comparing with the following video representations:

*1. Term attributes-f.* Similar to the few-example experiments in Section 5.1.1, this representation is extracted by the frequent term attributes, which are learned directly from terms without any embedding.

*2. VideoStory.* This video representation is learned by the VideoStory embeddings, proposed in Section 3.1, for few-example event recognition. More specifically, this representation is learned based on the LSI, in Eq. (2), as the descriptiveness loss, and the unimodal predictability, in Eq. (3), as the predictability loss.

**Results.** The results are reported in Table 5. The three embedding based representations, VideoStory, VideoStory$_0$, and VideoStory$_0^\mathcal{F}$, outperform the term attributes-f representation on both the MED 2013 and MED 2014 test sets. It confirms the effectiveness of embeddings for learning the semantic representations for zero-example event recognition. The most effective representation is learned by the VideoStory$_0^\mathcal{F}$, which rely on the term sensitive descriptiveness loss ($L_d^{\text{ts}}$), and the multimodal predictability loss ($L_p^\mathcal{F}$).

By comparing the term attributes-f and VideoStory, we observe that although the VideoStory utilizes the term correlations by embedding, its improvement over the term attributes-f is subtle, especially on the MED 2014 test set. As an explanation we highlight that in the VideoStory baseline, the descriptiveness loss is defined based on LSI. As discussed before, the LSI loss treats all the terms equally when measuring the reconstruction error. Hence this loss is biased toward minimizing the reconstruction error for the frequent terms. As a consequence, the VideoStory embedding generally is not accurate in predicting the terms, which are infrequent in its training data. It degrades the effectiveness of the VideoStory embedding for zero-example event recognition, where the accuracy in predicting some specific terms is crucial.

The drawback of the VideoStory baseline is addressed by VideoStory$_0$, which relies on the term sensitive descriptiveness loss. This loss minimizes the reconstruction error for the terms, which are indicative of the event, even if those terms are not frequent in the VideoStory train data. As a result, it can predict the distinctive terms of an event more accurately, compared to the VideoStory, which improves the event recognition accuracies from 15.9 to 18.3 on MED 2013, and from 5.2 to 6.8 on MED 2014. In Figure 9, we compare the term vectors, which are predicted by VideoStory and VideoStory$_0$ for three video examples. It shows that the VideoStory$_0$ more effectively predicts the indicative terms of the event, although they are not frequent in the train data.

The most effective representation is learned by the VideoStory$_0^\mathcal{F}$. It indicates that the term sensitive descriptiveness loss ($L_d^{\text{ts}}$) and the multimodal predictability loss ($L_p^\mathcal{F}$) are both effective and complementary when learning semantic video representations for zero-example event recognition.

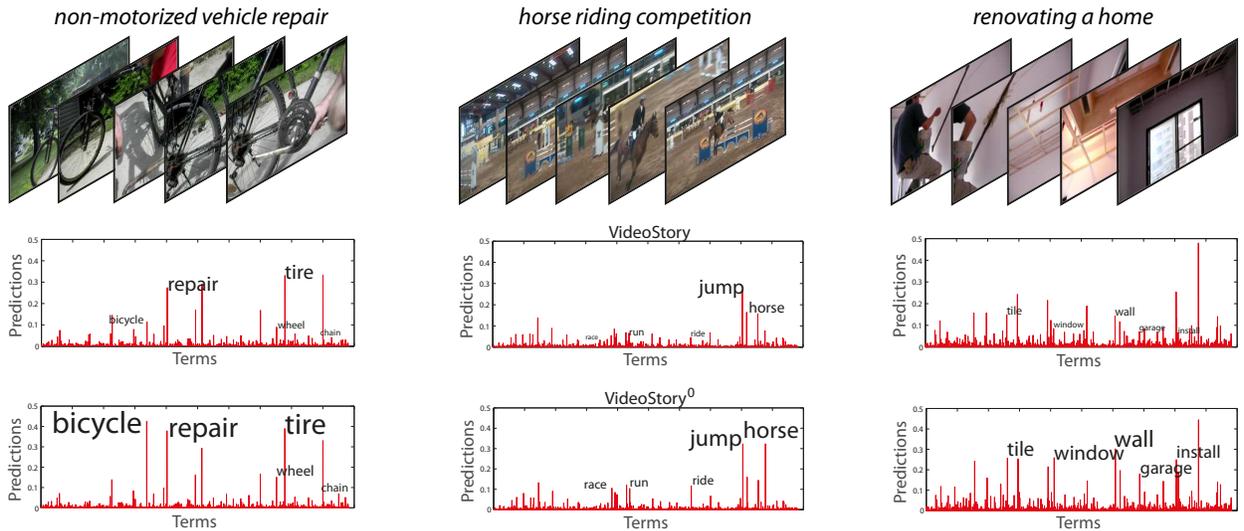

Fig. 9. Unseen video examples and their term vectors predicted by the VideoStory (middle) and the VideoStory$_0$ embeddings (bottom). The size of each term indicates its prediction confidence score. The VideoStory$_0$ more effectively predicts the indicative terms of the event.

TABLE 5
Effect of embedding for zero-example event recognition. The results demonstrate that the term sensitive descriptiveness loss ($L_d^{ts}$) and the multimodal predictability loss ($L_p^{\mathcal{F}}$) are both effective and complementary for zero-example event recognition.

| | MED 2013 | | | | | MED 2014 | | | |
|---|---|---|---|---|---|---|---|---|---|
| Event | Term attributes-f | VideoStory | VideoStory$_0$ | VideoStory$_0^{\mathcal{F}}$ | Event | Term attributes-f | VideoStory | VideoStory$_0$ | VideoStory$_0^{\mathcal{F}}$ |
| Birthday party | 14.5 | 24.6 | 30.3 | **37.4** | Attempting a bike trick | 4.9 | **8.8** | 7.1 | 5.9 |
| Changing a vehicle tire | 17.7 | **43.9** | 42.1 | 38.2 | Cleaning an appliance | 11.9 | 8.2 | 9.5 | **12.3** |
| Flash mob gathering | 15.2 | 14.5 | 22.4 | **33.8** | Dog show | **22.8** | 4.0 | 5.8 | 5.4 |
| Getting a vehicle unstuck | 28.9 | **40.2** | 36.0 | 32.2 | Giving directions to a location | 0.4 | 0.6 | 0.6 | **1.1** |
| Grooming an animal | 17.4 | 18.7 | **28.0** | 23.4 | Marriage proposal | 0.5 | 0.3 | 0.4 | **0.7** |
| Making a sandwich | 14.8 | 19.4 | **21.1** | 17.1 | Renovating a home | 4.8 | 5.2 | **6.8** | 6.3 |
| Parade | 20.0 | 17.6 | 26.4 | **38.2** | Rock climbing | 0.8 | **1.6** | 1.4 | 1.0 |
| Parkour | 15.9 | 26.1 | 28.0 | **40.3** | Town hall meeting | **15.5** | 1.9 | 9.2 | 9.2 |
| Repairing an appliance | 41.3 | 39.8 | 41.4 | **46.3** | Winning a race without a vehicle | 5.8 | **9.4** | 8.4 | 7.6 |
| Working on a sewing project | 23.8 | 30.8 | **36.4** | 36.0 | Working on a metal crafts | **7.5** | 1.5 | 4.7 | 4.8 |
| Attempting a bike trick | 5.0 | **8.8** | 7.1 | 5.9 | Beekeeping | 0.4 | 0.3 | 0.6 | **0.9** |
| Cleaning an appliance | 10.7 | 8.2 | 9.1 | **12.5** | Wedding shower | 1.3 | 2.3 | 3.3 | **3.3** |
| Dog show | **23.5** | 4.0 | 5.8 | 5.9 | Non-motorized vehicle repair | 8.4 | 33.2 | 44.2 | **46.9** |
| Giving directions to a location | 0.5 | 0.6 | 0.7 | **0.8** | Fixing musical instrument | 3.1 | 4.7 | 7.9 | **25.5** |
| Marriage proposal | 0.5 | 0.3 | 0.4 | **0.7** | Horse riding competition | 2.3 | 7.7 | 12.8 | **13.7** |
| Renovating a home | 4.8 | 5.2 | **6.8** | 6.4 | Felling a tree | 3.9 | **7.7** | 6.5 | 6.8 |
| Rock climbing | 0.8 | **1.6** | 1.4 | 1.0 | Parking a vehicle | 0.3 | 2.8 | 2.3 | **2.9** |
| Town hall meeting | **15.5** | 1.9 | 9.2 | 9.6 | Playing fetch | **3.6** | 2.2 | 2.9 | 3.3 |
| Winning a race without a vehicle | 5.8 | **9.4** | 8.4 | 7.9 | Tailgating | **0.6** | 0.2 | 0.2 | 0.4 |
| Working on a metal crafts | **7.3** | 1.6 | 4.9 | 7.0 | Tuning musical instrument | 0.9 | 0.5 | 0.9 | **2.0** |
| mAP | 14.2 | 15.9 | 18.3 | **20.0** | mAP | 5.0 | 5.2 | 6.8 | **8.0** |

### 5.4 Comparison with the state-of-the-art

We evaluate the merit of the proposed VideoStory by comparing it with several other recent works on few- and zero-example event recognition. Since in most papers the results are reported only on the MED 2013 test set, we limit our comparisons to this test set.

The results are reported in Table 6. It shows that our proposed representation learning sets a new state-of-the-art for the both few- and zero-example event recognition. It should be noted that these works rely on modeling the temporal aspects of the events [8], [40], [63], [66], using larger train set for representation learning [7], [29], [31], and query expansion [7], [76] to improve the event recognition. These improvements can be also applied together with our representation learning for a more effective event recognition. Very recently Jiang *et al.* [31] improved their results for zero-example recognition from 18.3 to 20.8 after adding re-ranking by pseudo relevance feedback, we expect a similar gain for VideoStory$_0$ and VideoStory$_0^{\mathcal{F}}$.

## 6 CONCLUSION

In this paper, we attack the problem of event recognition in video when examples are scarce. We propose the *VideoStory* embedding that learns a semantic video representation from a set of videos and their textual descriptions by minimizing a joint objective function balancing term descriptiveness and video predictability losses. As a result, the terms which are correlated in the descriptions are combined together to improve their video predictability. Besides its effectiveness for few-example event recognition, the VideoStory embedding also improves the representation learning time by reducing redundancies during training.

In addition, we propose the *VideoStory$^{\mathcal{F}}$* embedding with a multimodal predictability loss learned jointly over video appearance, motion, and audio features. The different modalities are complementary for predicting the terms, so they are all required for a richer video description. Moreover, embedding the heterogeneous video features into a mutual semantic space leads to few-example event recogni-tion that is more effective than traditional fusion tactics.

TABLE 6
Comparison with the state-of-the-art. VideoStory embeddings set a new state-of-the-art for the both few- and zero-example event recognition.

| Few-example on MED 2013 | | | Zero-example on MED 2013 | | |
|---|---|---|---|---|---|
| Habibian et al. [20] | MM 2014 | 19.6 | Ye et al. [80] | MM 2015 | 9.0 |
| Nagel et al. [48] | BMVC 2015 | 21.8 | Chang et al. [7] | IJCAI 2015 | 9.6 |
| Li et al. [40]† | ICCV 2013 | 23.7 | Mazloom et al. [46] | ICMR 2015 | 11.9 |
| Tang et al. [66]† | CVPR 2012 | 26.8 | Wu et al. [76] | CVPR 2014 | 12.7 |
| Sun et al. [63]† | CVPR 2014 | 28.7 | Jiang et al. [29] | AAAI 2015 | 12.9 |
| Chang et al. [8] | MM 2015 | 30.9 | Jiang et al. [31] | MM 2015 | 18.3 |
| This paper: *VideoStory* | | 32.4 | This paper: *VideoStory$_0$* | | 18.3 |
| This paper: *VideoStory$^\mathcal{F}$* | | **37.1** | This paper: *VideoStory$_0^\mathcal{F}$* | | **20.0** |

† Based on implementation by [8]

We also propose the *VideoStory$_0$* embedding to recognize an event in video from just a textual query in the form of an event description. This embedding relies on a term sensitive descriptiveness loss to learn a more accurate representation for the terms, which are indicative of the event. Finally, the *VideoStory$_0^\mathcal{F}$* embedding demonstrates that the term sensitive descriptiveness loss and the multimodal predictability loss are both effective and complementary to learn semantic video representations for zero-example event recognition.

By it abilities to improve predictability upon any underlying video feature while at the same time maximizing semantic descriptiveness, VideoStory leads to state-of-the-art accuracy for both few- and zero-example recognition of events in video. We consider VideoStory's ability to generate human interpretable representations for previously unseen videos most appealing, as it opens up new connections with natural language processing and computational linguistics for describing and querying videos.